# Monocular Depth Estimation with Global-Aware Discretization and Local Context Modeling

Heng Wu, Qian Zhang[*], Guixu Zhang

*School of Computer Science and Technology, East China Normal University, Shanghai, China*

**Abstract**

Accurate monocular depth estimation remains a challenging problem due to the inherent ambiguity that stems from the ill-posed nature of recovering 3D structure from a single view, where multiple plausible depth configurations can produce identical 2D projections. In this paper, we present a novel depth estimation method that combines both local and global cues to improve prediction accuracy. Specifically, we propose the Gated Large Kernel Attention Module (GLKAM) to effectively capture multi-scale local structural information by leveraging large kernel convolutions with a gated mechanism. To further enhance the global perception of the network, we introduce the Global Bin Prediction Module (GBPM), which estimates the global distribution of depth bins and provides structural guidance for depth regression. Extensive experiments on the NYU-V2 and KITTI dataset demonstrate that our method achieves competitive performance and outperforms existing approaches, validating the effectiveness of each proposed component.

## 1. Introduction

Depth estimation endeavors to infer the distance between scene points and the camera based on a single RGB image, providing critical geometric understanding for various computer vision applications, such as autonomous driving [43], robot navigation [4], and 3D scene reconstruction [5]. However, monocular depth estimation (MDE) remains fundamentally challenging task due to the lack of geometric priors such as stereo correspondences or temporal cues.

Convolutional neural network (CNN)-based methods have played a leading role in the early development of this field, benefiting from their deep architectural designs, such as DCNN [8], ResNet [12], MobileXNet [6], and EfficientNet [2], to extract powerful local features. Although these models effectively capture fine-grained information, their inherent locality—resulting from limited receptive fields—restricts their ability to model global scene structures, which play a crucial role in achieving accurate depth estimation within complex environments. To address the limitations of CNNs, Transformer-based architectures have recently been introduced into the monocular depth estimation (MDE) task. Models such as Vision Transformer (ViT) [30] and SwinTransformer [1,22] have demonstrated remarkable capabilities in capturing long-range dependencies, benefiting from their global receptive fields. Furthermore, hybrid approaches that combine the strengths of CNNs and Transformers have been proposed to leverage both local detail modeling and global contextual reasoning simultaneously. For example, DepthFormer [21] integrates convolutional and Transformer modules to enhance feature representations and achieve promising results.

Despite the success of recent deep learning-based methods, current approaches still suffer from three major limitations. First, many networks rely on local convolutional operations or windowed attention mechanisms, which resulting in limited insufficient receptive fields that hinder the modeling of long-range spatial dependencies. Second, the ability to capture multi-scale contextual information is often limited, making it difficult to accurately estimate depth across objects of varying sizes and across different scene scales. Third, existing depth regression strategies typically discretize depth values into bins using fixed or locally adaptive heuristics, which lack global awareness and often lead to suboptimal predictions in complex environments. These challenges significantly limit the accuracy and generalization ability of current MDE systems, particularly in scenes with diverse geometries and wide depth ranges.

In response to these challenges, we propose a novel monocular depth estimation method featuring two key innovations: a Gated Large Kernel Attention Module (GLKAM) and a Global Bin Prediction Module (GBPM). Specifically, the GLKAM introduces large convolutional kernels in a multi-scale design to expand the effective receptive field, enabling the network to capture both local details and global structures more comprehensively. Meanwhile, the GBPM rethinks the traditional binning approach by predicting global depth distribution bins, thereby facilitating long-range depth reasoning and enhancing depth prediction precision.

We validate the effectiveness of the proposed method through extensive experiments conducted on two widely used benchmarks: NYU-V2 and KITTI. The results demonstrate that our method achieves superior performance across multiple evaluation metrics, particularly excelling in complex scenes characterized by diverse geometries and varying depth distributions.

The main contributions of this work can be summarized as follows:
- We propose a Gated Large Kernel Attention Module (GLKAM), which efficiently captures multi-scale contextual features and significantly expands the receptive field.
- We design a Global Bin Prediction Module (GBPM) that predicts depth bins under a global context, improving the precision of depth regression.
- We validate the proposed method through extensive quantitative and qualitative experiments on NYU-V2 and KITTI datasets, outperforming recent methods and demonstrating the effectiveness of our method.

The remaining sections of the paper are organized as follows: Section 2 reviews related work, Section 3 provides a detailed explanation of the proposed method, Section 4 presents the experimental results, and Section 5 concludes the paper.

## 2. Related Work

Monocular depth estimation (MDE) has witnessed significant advancements due to the rapid development of deep learning techniques. Existing approaches can be mainly categorized into CNN-based methods and Transformer-based methods. In addition, hybrid Transformer-CNN architectures have recently drawn considerable attention. In this section, we provide a detailed review of these three categories.

### 2.1. CNN-based MDE methods

The pioneering work by Eigen et al. [7] introduced the first CNN-based framework for monocular depth estimation, employing a coarse-to-fine prediction strategy. Subsequent research further explored the potential of CNNs in MDE by designing deeper networks and incorporating advanced convolutional operations. Fu et al. [8] proposed a deep ordinal regression network that discretized depth values into intervals and modeled ordinal relationships. Lee et al. [17] introduced local planar guidance to enhance the up-sampling process in the decoder. Multi-scale feature fusion has also been widely adopted to aggregate information across different resolutions [3,20,23]. To better capture spatial dependencies, Huynh et al. [14] and Liu et al. [23] integrated attention mechanisms into CNN frameworks. Despite their impressive performance, CNN-based methods often suffer from limited receptive fields, constraining their ability to model long-range dependencies critical for global scene understanding.

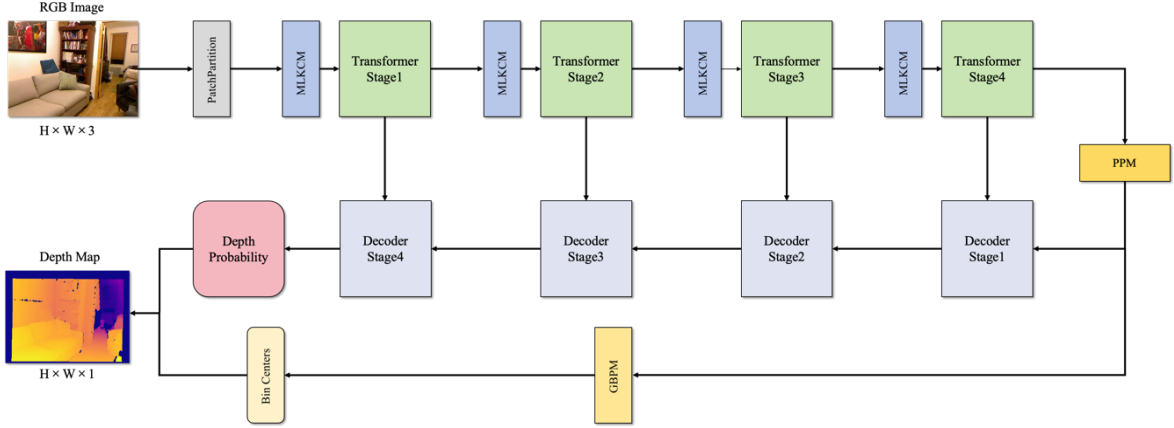

**Figure 1**: Overview of the proposed network architecture. The framework follows an encoder-decoder paradigm based on a SwinTransformer backbone. A series of GLKAM are inserted between encoder stages to enhance multi-scale local feature extraction. Decoder stages progressively fuse multi-scale features with skip connections, and a PPM is used for global context aggregation. Finally, the GBPM predicts global bin centers and per-pixel depth distributions to produce the final depth map.

*2.2. Transformer-based MDE methods*

With the advent of Vision Transformers (ViTs) [36], researchers have leveraged their strong global modeling capabilities for MDE tasks. Ranftl et al. [30] proposed a Transformer backbone for dense prediction tasks, including depth estimation. Bhat et al. [2] introduced AdaBins, which adaptively partitioned the depth range into bins using a Transformer refinement network. Yuan et al. [42] presented a neural window fully connected CRF decoder combined with ViT features. Selective feature fusion strategies were explored by Kim et al. [15] to enhance depth map generation. Furthermore, several works incorporated adaptive binning strategies based on the SwinTransformer backbone to boost depth estimation accuracy [22,32]. Ning and Gan [27] proposed a novel trap attention mechanism for depth regression. Although Transformer-based methods excel at capturing global context, their relatively high computational cost and less effective local feature extraction remain challenges.

*2.3. The hybrid methods*

Recognizing the complementary strengths of CNNs and Transformers, hybrid architectures have been developed to better balance local detail preservation and global structure modeling. One strategy is to integrate CNN operations into Transformer blocks. For instance, MPViT [18] embeds convolutional paths within Transformer layers, while Guo et al. [11] propose a hybrid convolution-Transformer model for efficient feature extraction. Another approach involves parallel CNN and Transformer branches, allowing each to specialize in local or global information. Li et al. [21] adopted this idea with a hierarchical aggregation module. Furthermore, recent studies have demonstrated that hybrid designs can improve performance across various dense prediction tasks, including depth estimation [31,40,44]. For example, Zhang et al. [44] proposed a lightweight depth estimation network coupling CNNs and Transformers, and Shao et al. [31] introduced an uncertainty-rectified cross-distillation framework.

These methods underline the effectiveness of hybrid designs in enhancing depth prediction accuracy.

## 3. Proposed Method

This section firstly presents the overall architecture of our network. A detailed description of the proposed GLKAM and GBPM will be provided in subsequent sections. Finally, we specify our loss function.

### 3.1. Overall Architecture

The entire architecture is illustrated in **Figure 1**, our proposed method is designed to improve depth estimation performance by integrating both global and local contextual information. The overall framework comprises three key components: a backbone encoder-decoder for multi-scale feature extraction, a Gated Large Kernel Attention Module (GLKAM) for adaptive spatial context modeling, and a Global Bin Prediction Module (GBPM) for adaptive depth distribution prediction. We use the SwinTransformer backbone with a four-stage pyramid structure as the encoder for processing RGB images. Given an input RGB image with resolution H×W, the outputs of the four stages are the encoder features $\{E_1, E_2, E_3, E_4\}$ at scales $\{1/4, 1/8, 1/16, 1/32\}$ of original image respectively, with channel dimensions $\{C_1, C_2, C_3, C_4\} = \{C, 2C, 4C, 8C\}$, where C is set to 192 as per the configuration in [24]. In the encoding stage, the Gated Large Kernel Attention Module depicted in **Figure 2** is used between each two adjacent stages to extract multi-scale local features as a supplement to the prepositional local information of the SwinTransformer stages.

After the four-stage encoding process, we employ a Pyramid Pooling Module (PPM) [45] to further enhance global context aggregation. The PPM aggregates features from multiple spatial scales by applying pooling operations of different kernel sizes, followed by 1 × 1 convolutions to project them into a unified feature space. These multi-scale features are then up-sampled and fused, effectively capturing long-range dependencies and providing complementary information to the decoder features.

The PPM output is sent to a Global Bin Prediction Module (GBPM) depicted in **Figure 3**, it generates a globally optimized bin distribution and bin centers, allowing the network to better handle depth variations across diverse scenes.

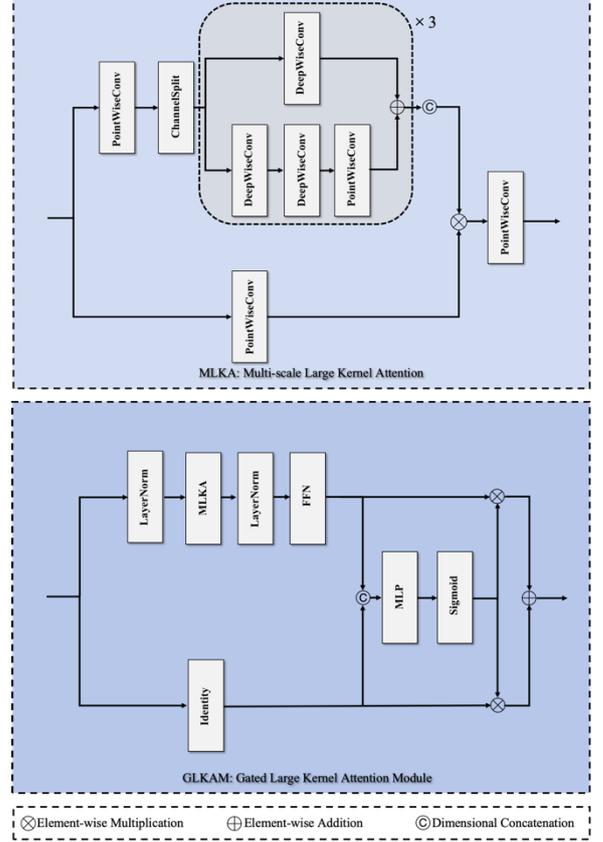

**Figure 2**: Details of the proposed Gated Large Kernel Attention Module. This module combines multi-scale depth-wise convolutions with a dynamic gating mechanism. It extracts rich local context through grouped large kernel convolutions and modulates features via an attention path consisting of MLKA, FFN, and a Sigmoid-based gating branch.

The decoder reconstructs depth feature maps by progressively up-sampling and fusing multi-scale encoder features. It consists of four hierarchical stages, each corresponding to one encoder output. The outputs of the four stages are the decoder features $\{D_1, D_2, D_3, D_4\}$ at scales $\{1/32, 1/16, 1/8, 1/4\}$ of original image respectively. In each stage, the corresponding encoder feature $E_i$ and decoder feature $D_i$ are concatenated firstly and up-sampled via pixel shuffle for feature refinement. The final decoder stage feature $D_4$ is used to generate the depth distribution probability.

*3.2. Gated Large Kernel Attention Module*

To effectively capture multi-scale local features and enhance spatial detail representation, we propose the Gated Large Kernel Attention Module (GLKAM), which serves as a critical component before feeding features into the Transformer-based global modeling stage.

In depth estimation, local structural cues such as object boundaries, surface textures, and geometric patterns play a crucial role in resolving depth ambiguities. However, convolutional layers with limited kernel sizes are insufficient to model rich spatial dependencies, while naively stacking deep convolutional layers increases computational cost and risks overfitting. To address these limitations, GLKAM introduces a multi-scale large kernel attention mechanism that balances large receptive fields with computational efficiency. As shown in **Figure 2**, given the input feature map $X \in R^{C \times H \times W}$, we first normalize X and pass through a Multi-scale Large Kernel Attention (MLKA) [37] module to generate a multi-scale local context-enhanced feature. To begin with, the input feature map undergoes a point-wise convolution ($1 \times 1$) to align the number of channels as needed. The transformed feature tensor is then divided into three channel-wise segments via a Channel Split operation. Each of these segments is processed independently by a Large Kernel Attention (LKA) module, each configured with distinct receptive fields. For i-th group of features $X_i$, $i \in \{1, 2, 3\}$, an LKA decomposed by $\{K_i, d_i\}$ is utilized to generate a homogeneous scale attention map $LKA_i$. In detail, LKA is leveraged into three groups: $\{7, 2\}$ implemented by 3-5-1, $\{21, 3\}$ by 5-7-1, and $\{35, 4\}$ by 7-9-1, where a-b-1 means cascading a × a depth-wise, b × b depth-wise-dilated, and point-wise convolutions. To enable adaptive control over the attention response, each LKA module is modulated by a spatial gate. The gated output of each path is formulated as:

$$MLKA_i(X_i) = G_i(X_i) \otimes LAK_i(X_i) \quad (1)$$

where $G_i(.)$ is a spatial gate generated via a depth-wise convolution of size $a_i \times a_i$, and $LKA_i(.)$ is the attention map constructed by the corresponding kernel sequence $a_i$-$b_i$-1. The MLKA branch can be formulated as:

$$F_m = FFN\left(LN\left(MLKA(LN(X))\right)\right) \quad (2)$$

To combine the attentive features with the original representation, we concatenate the original input and the attended feature map. A lightweight MLP followed by a sigmoid activation is used to learn a soft gate:

$$F = CONCAT(F_m, X) \quad (3)$$
$$z = SIGMOID(MLP(F)) \quad (4)$$

This gate dynamically controls the fusion of the original and attentive features at each spatial location and channel.

The final output $F_{out}$ is a gated combination of the original input and the enhanced feature:

$$F_{out} = z \otimes X + (1 - z) \otimes F_m \quad (5)$$

After extracting the local multi-scale feature $F_{out}$, it is sent to the SwinTransformer stage.

*3.3. Global Bin Prediction Module*

The Global Bin Prediction Module (GBPM) is designed to explicitly predict the depth bin distribution of the entire scene, providing global structural guidance for dense depth estimation. Unlike uniform binning strategies that divide the depth range into fixed intervals [8], GBPM dynamically determines the positions of depth bins based on the global context of the input image, allowing the network to adaptively allocate representation capacity according to scene complexity.

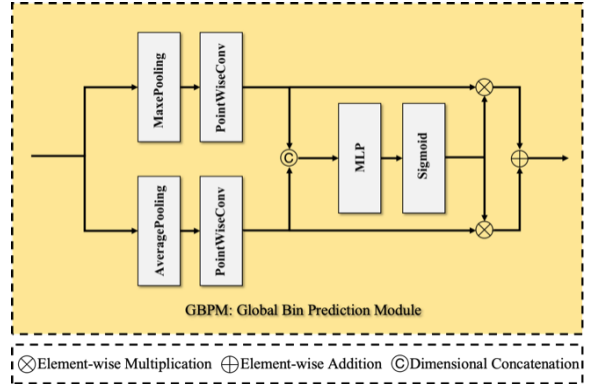

**Figure 3**: Details of the proposed Global Bin Prediction Module. This module aggregates global contextual cues using parallel max and average pooling branches followed by point-wise convolutions. The outputs are concatenated and passed through an MLP and Sigmoid to generate gate weights, which are used to adaptively recalibrate the input features for robust global depth bin prediction.

As shown in **Figure 3**, given the multi-scale feature map extracted by the encoder and PPM, denoted as $F \in R^{C \times H \times W}$. First, we extract a global semantic descriptor $F_a$ by applying global average pooling (GAP) followed by a point-wise convolution (PW Conv) to the input feature map F. Simultaneously, we apply a global max pooling (GMP) operation to capture prominent geometric structures feature $F_b$ in the scene, followed by another point-wise convolution:

$$F_a = PWConv(GAP(F)) \quad (6)$$
$$F_b = PWConv(GMP(F)) \quad (7)$$

The outputs from the two branches are concatenated and passed through a lightweight multi-layer perceptron (MLP), followed by a sigmoid activation to compute the gating coefficient z. Then, the two global features are fused via weighted summation based on the learned gate.

$$z = SIGMOID(MLP(CONCAT(F_a, F_b))) \quad (8)$$
$$F_{out} = F_a \otimes z + F_b \otimes (1-z) \quad (9)$$

Finally, the bin width b is predicted by a MLP layer and bin centers for the input image are computed as:

$$b = MLP(F_{out}) \quad (10)$$
$$c(b_i) = d_{min} + (d_{max} - d_{min})\left(\frac{b_i}{2} + \sum_{j=1}^{i-1} b_j\right) \quad (11)$$

### 3.4. Depth Prediction

The final depth is predicted by the linear combination of the bin centers weighted by the depth probability generated by the decoder:

$$d_i = \sum_{k=1}^{n_{bins}} c(b_k) p_{ik} \quad (12)$$

where $d_i$ is the predicted depth at pixel i, $c(b_k)$ is the k-th bin center, $n_{bins}$ are the number of bins, and $p_{ik}$ is the probability for bin center k for a pixel i. We set $n_{bins}$ = 256 according to [2].

### 3.5. Loss Function

Following previous works [2], we adopt a modified version of the Scale-Invariant Logarithmic (SILog) loss [7] as the supervision objective for training our network.

$$L_{SILog} = \alpha\sqrt{\frac{1}{n}\sum_i g_i^2 - \frac{\lambda}{n}\left(\sum_i g_i\right)^2} \quad (13)$$

Where $g_i = log(d_i^*) - log(d_i)$ and $d_i$ is the ground truth depth value of pixel i, $d_i$, n is the total valid pixel number in image. Here is set as 0.85 and is set as 10 for all our experiments.

## 4. Experimental Result

In this section, we conducted experimental comparisons to evaluate the performance of the proposed method in comparison to previous methods. Additionally, we analyzed the impact of the proposed GLKAM and GBPM through ablation studies.

### 4.1. Experimental setup

**Datasets.** We evaluate our proposed method on two widely used benchmark datasets: NYU-V2 and KITTI. NYU-V2 [33] is an indoor dataset collected using a Microsoft Kinect sensor, comprising approximately 120K RGB-depth image pairs captured from 464 different indoor scenes. Each image has a spatial resolution of 480 × 640. Following the standard experimental protocol, we adopt the official split: 50K images from 249 scenes for training, and 654 images from 215 scenes for evaluation. Consistent with Eigen et al. [7], we apply center cropping and limit the maximum depth value to 10 meters.

KITTI [10] is a large-scale outdoor dataset acquired from real-world urban driving scenarios. It provides stereo RGB images and corresponding sparse depth maps obtained from LIDAR sensors, captured across 61 scenes using a multi-sensor platform mounted on a moving vehicle. Each RGB image has a resolution of 1241 × 375 pixels. We adopt the training and testing splits introduced by Eigen et al. [7], using 26K left-view images for training and 697 images for evaluation. During testing, we apply the cropping strategy defined by Garg et al. [9], and restrict depth values to a maximum of 80 meters. The predicted depth maps are bilinearly up-sampled to match the original resolution before computing the evaluation metrics.

**Implementation details.** Our method is developed using the PyTorch framework [28]. We adopt the Adam optimizer [16] with default momentum parameters ($\beta_1$ = 0.9, $\beta_2$ = 0.999), and set the batch size to 8 along with a weight decay of 0.01. For both NYU-V2 and KITTI datasets, we train the model for 20 epochs. The learning rate is initialized at $4\times10^{-5}$ and linearly reduced to 4×10⁻⁶ over the course of training. All experiments are conducted on a server equipped with four NVIDIA RTX 3090 GPUs.

To improve generalization, we apply several data augmentation strategies during training, including random rotations, horizontal flips, and variations in image brightness. The Swin-L backbone is initialized with ImageNet pre-trained weights provided in [24]. During inference, we adopt the common practice used in [2,42], where predictions are generated for both the input image and its horizontally flipped version, and the final depth map is obtained by averaging the two results.

**Evaluation Metrics.** We employed standard error and accuracy evaluation metrics commonly utilized in MDE studies [2,7,22] for quantitative assessment: Average relative error (Abs Rel), Root mean squared error (RMSE), Average Log error (log10), Threshold Accuracy(i) at thresholds 1.25, $1.25^2$, $1.25^3$. Given the predicted depth $d_{pred}$, and the ground truth depth $d_{gt}$, and n denoting the total number of pixels in an image, the error evaluation metrics are defined as follows.

$$AbsRel = \frac{1}{n}\sum \frac{|d_{pred} - d_{gt}|}{d_{gt}} \quad (14)$$

$$Log10 = \frac{1}{n}\sum |d_{pred} - d_{gt}| \quad (15)$$

$$RMSE = \sqrt{\frac{1}{n}\sum (d_{pred} - d_{gt})^2} \quad (16)$$

$$SqRel = \frac{1}{n}\sum \frac{\|d_{pred} - d_{gt}\|^2}{d_{gt}} \quad (17)$$

$$\delta_i = max\left(\frac{d_{pred}}{d_{gt}}, \frac{d_{gt}}{d_{pred}}\right) < 1.25^i, i \in \{1,2,3\} \quad (18)$$

## 4.2. Performance Comparison

In this section, we will introduce the qualitative and quantitative results of the proposed method on the NYU-V2 and KITTI datasets.

### 4.2.1. Experiment on NYU-V2

**Table 1** presents the experimental results of our method on the NYU-V2 dataset. The results showed that our method achieved three accuracy metrics of 0.934, 0.992, and 0.999, all of which were higher than other methods, indicating a significant improvement in the accuracy of depth estimation. Our method demonstrates a significant improvement of 6.3%, 4.8%, 5% in the AbsRel metric, RMSE metric and log10 metric compared to the BinsFormer method.

As illustrated in **Figure 4**, our method demonstrates superior qualitative performance, particularly inaccurately delineating local geometric structures such as table edges, chair contours, and other indoor objects. Compared with prior approaches, our predictions exhibit clearer depth boundaries and more continuous transitions in ambiguous regions. This improvement is largely attributed to the GLKAM, which effectively captures rich multi-scale local context by expanding the receptive field and selectively enhancing informative features.

In addition to refined local details, our method also better preserves global scene structures and maintains coherent depth gradients across large spatial regions. This is primarily enabled by the GBPM, which globally estimates bin centers, allowing the network to model scene-level depth distributions more accurately. Together, the GLKAM and GBPM modules contribute to the model's ability to integrate fine-grained local cues with holistic global understanding, resulting in more reliable and structurally consistent depth predictions. These qualitative observations are consistent with our quantitative results, confirming the effectiveness of our design for indoor monocular depth estimation.

### 4.2.2. Experiment on KITTI

**Table 2** presents the experimental results of our method on the KITTI dataset. The results show that our method achieved three accuracy metrics of 0.977, 0.997, and 0.999, all of which are higher than other methods, indicating a significant improvement in the accuracy of depth estimation. Our method demonstrates a significant improvement of 3.8%, 2.5%, 1.9% in the AbsRel metric, log10 metric and SqRel metric compared to the BinsFormer method.

**Figure 5** presents qualitative comparison results on the KITTI dataset, our method produces more accurate and visually consistent depth maps, especially in challenging outdoor scenes characterized by occlusions, shadows, and varying illumination. The improvements are most evident in the regions marked by dotted boxes. For example, our method recovers finer object boundaries such as poles, traffic signs, and distant cars, which are often blurred or distorted in the predictions of DepthFormer and BinsFormer.

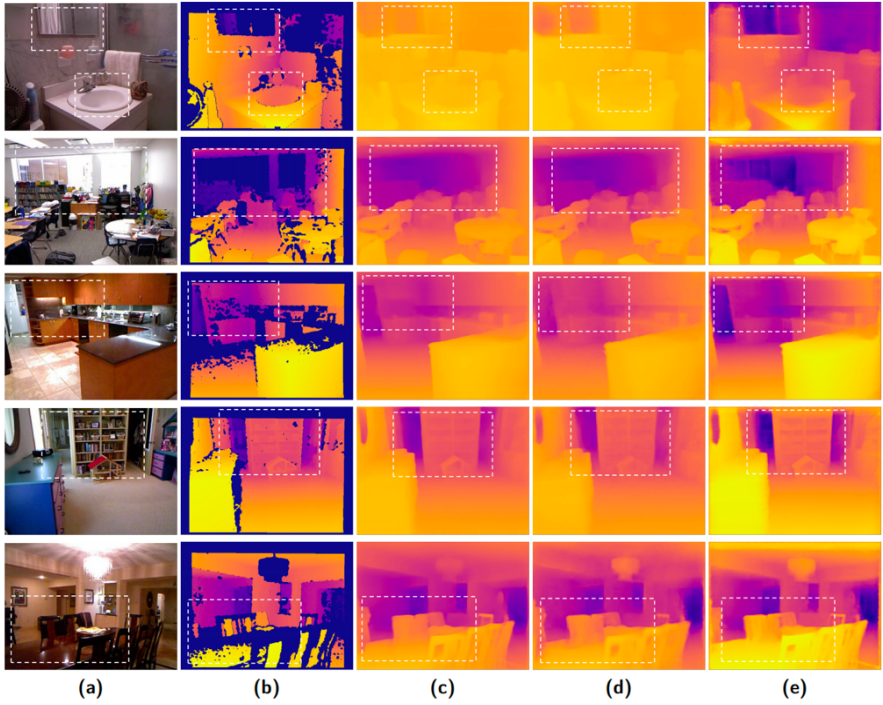

**Figure 4**: Qualitative comparison results on NYU-V2. Color indicates depth (red is close, blue is far). (a) RGB images, (b) ground truth, (c) DepthFormer [21], (d) BinsFormer [22], (e) ours.

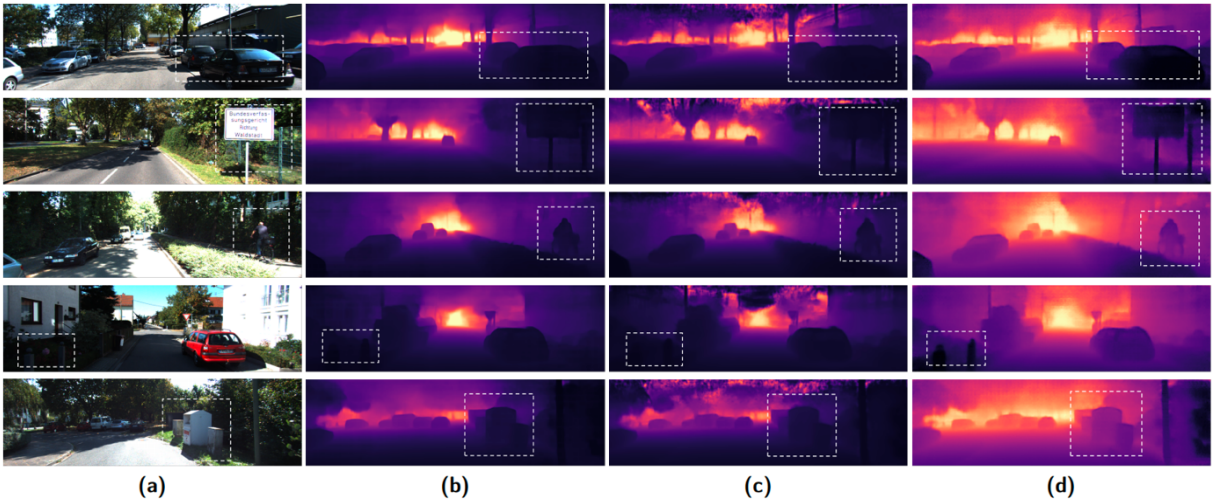

**Figure 5**: Qualitative comparison results on KITTI. Color indicates depth (red is far, blue is close). (a) RGB images, (b) DepthFormer [21], (c) BinsFormer [22], (d) ours.

In addition, our method demonstrates superior depth continuity in large homogeneous regions such as roads and building facades. This is attributed to the combination of the Gated Large Kernel Attention

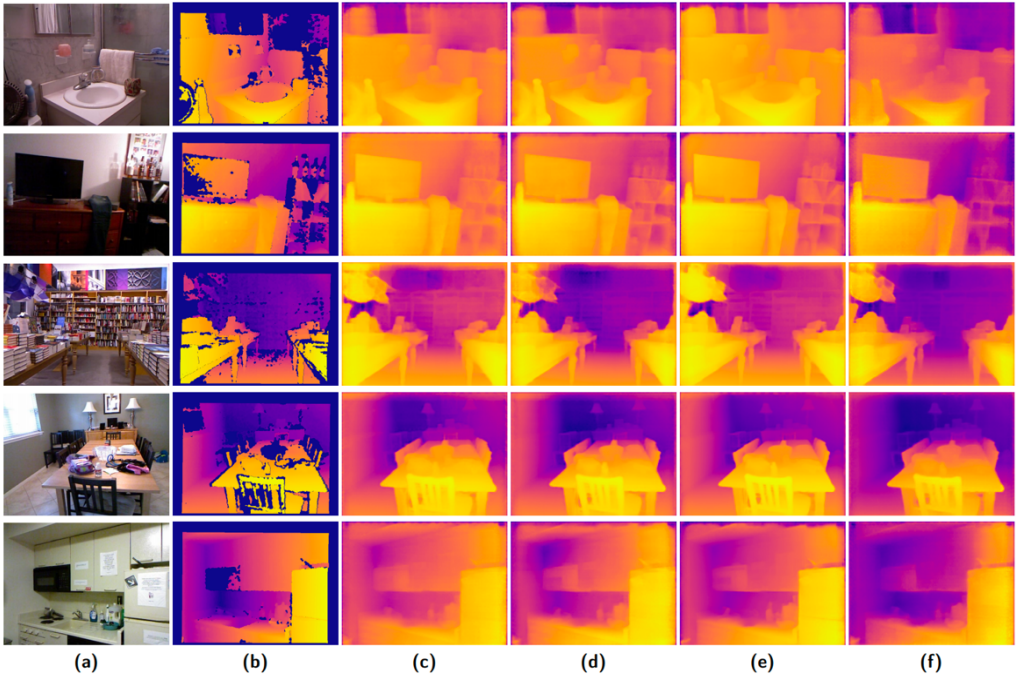

**Figure 6**: Ablation study on NYU V2 dataset. Color indicates depth (red is close, blue is far). (a) RGB images, (b) ground truth, (c) baseline, (d) baseline+GBPM, (e) baseline+GLKAM, (f) baseline+GLKAM+GBPM.

Module (GLKAM), which enhances spatial detail recovery, and the Global Bin Prediction Module (GBPM), which improves global depth reasoning.

### 4.3. Ablation Study

To better understand the effectiveness of each proposed component, we conduct comprehensive ablation experiments on the NYU-V2 dataset. As shown in **Table 3**, we evaluate the individual and combined impact of the GLKAM and the GBPM on the final performance.

#### 4.3.1. Effectiveness of GLKAM

As shown in **Table 3**, the inclusion of GLKAM yields consistent improvements across all evaluation metrics. Specifically, the $\delta_1$ accuracy increases from 0.924 to 0.932, and RMSE drops from 0.322 to 0.316. These results indicate that GLKAM effectively enhances feature representation by capturing richer local context, leading to more accurate depth predictions. In particular, the reduction in both AbsRel and $\log_{10}$ suggests improved pixel-level accuracy and overall depth consistency.

**Table 1**: Quantitative results on NYU-V2 Dataset. The best scores are highlighted in bold and second best are underlined. ↑ means higher the better and ↓ means lower the better.

| Method | $\delta_1$↑ | $\delta_2$↑ | $\delta_3$↑ | AbsRel↓ | RMSE↓ | $\log_{10}$↓ |
|---|---|---|---|---|---|---|
| Eigen et al. [7] | 0.769 | 0.950 | 0.988 | 0.158 | 0.641 | - |
| DORN[8] | 0.828 | 0.965 | 0.992 | 0.115 | 0.509 | 0.051 |
| Yin et al. [41] | 0.875 | 0.976 | 0.994 | 0.108 | 0.416 | 0.048 |
| BTS[17] | 0.885 | 0.978 | 0.994 | 0.110 | 0.392 | 0.047 |
| Xiao et al. [38] | 0.895 | 0.983 | 0.996 | 0.104 | 0.380 | 0.045 |
| P3Depth[29] | 0.898 | 0.981 | 0.996 | 0.104 | 0.356 | 0.043 |
| Li et al. [19] | 0.898 | 0.981 | 0.996 | 0.104 | 0.356 | **0.038** |
| TransDepth[39] | 0.900 | 0.983 | 0.996 | 0.106 | 0.365 | 0.045 |
| AdaBins[2] | 0.903 | 0.984 | 0.997 | 0.103 | 0.364 | 0.044 |
| DPT[30] | 0.904 | 0.988 | **0.998** | 0.110 | 0.367 | 0.045 |
| CATNet[35] | 0.915 | <u>0.989</u> | <u>0.997</u> | <u>0.093</u> | 0.338 | <u>0.040</u> |
| NeWCRFs[42] | 0.922 | **0.992** | 0.998 | 0.095 | 0.334 | 0.041 |
| DepthFormer[21] | 0.923 | <u>0.989</u> | 0.997 | <u>0.094</u> | 0.329 | <u>0.040</u> |
| BinsFormer[22] | <u>0.925</u> | <u>0.989</u> | 0.997 | <u>0.094</u> | 0.330 | <u>0.040</u> |
| Our Method | **0.934** | **0.992** | **0.998** | **0.088** | **0.314** | **0.038** |

**Figure 6** further demonstrates the visual benefits of GLKAM. Compared to the baseline, the model with GLKAM produces sharper depth boundaries and clearer object structures, such as sinks, chairs and shelves. The predicted depth maps exhibit better preservation of local geometric details, particularly in cluttered or high-frequency regions. These improvements can be attributed to GLKAM's ability to aggregate multi-scale local features using large-kernel convolutions and dynamically modulate them via its gating mechanism.

Table 2: Quantitative results on KITTI Dataset. The best scores are highlighted in bold and second best are underlined. ↑ means higher the better and ↓ means lower the better.

| Method | $\delta_1$ ↑ | $\delta_2$ ↑ | $\delta_3$ ↑ | AbsRel↓ | SqRel↓ | $\log_{10}$ ↓ |
|---|---|---|---|---|---|---|
| Eigen et al. [7] | 0.702 | 0.898 | 0.967 | 0.203 | 1.548 | 0.282 |
| GCNDepth[25] | 0.888 | 0.965 | 0.984 | 0.104 | 0.720 | 0.181 |
| RA-Depth[13] | 0.903 | 0.968 | 0.985 | 0.096 | 0.632 | 0.171 |
| SDF-Net[46] | 0.907 | 0.969 | 0.985 | 0.089 | 0.531 | 0.168 |
| DORN[8] | 0.932 | 0.984 | 0.994 | 0.072 | 0.307 | 0.120 |
| Naderi et al. [26] | 0.944 | 0.991 | <u>0.998</u> | 0.070 | - | 0.113 |
| TransDepth[39] | 0.956 | 0.994 | **0.999** | 0.064 | 0.252 | 0.098 |
| BTS[17] | 0.956 | 0.993 | <u>0.998</u> | 0.059 | 0.245 | 0.096 |
| DPT[30]] | 0.959 | <u>0.995</u> | **0.999** | 0.062 | - | 0.092 |
| LapDepth[34] | 0.962 | 0.994 | **0.999** | 0.059 | 0.212 | 0.091 |
| AdaBins[2] | 0.964 | <u>0.995</u> | **0.999** | 0.058 | 0.190 | 0.088 |
| CATNet[35]] | 0.972 | **0.997** | **0.999** | 0.052 | 0.160 | 0.081 |
| NeWCRFs[42] | 0.974 | **0.997** | **0.999** | <u>0.052</u> | 0.155 | <u>0.079</u> |
| DepthFormer[21] | <u>0.975</u> | **0.997** | **0.999** | 0.052 | 0.158 | <u>0.079</u> |
| BinsFormer[22] | 0.974 | **0.997** | **0.999** | <u>0.052</u> | <u>0.151</u> | <u>0.079</u> |
| Our Method | **0.977** | **0.997** | **0.999** | **0.050** | **0.148** | **0.077** |

Table 3: Ablation study on NYU-V2 dataset. ↑ means higher the better and ↓ means lower the better. The best scores are highlighted in bold.

| Setting | $\delta_1$ ↑ | AbsRel↓ | RMSE↓ | $\log_{10}$ ↓ |
|---|---|---|---|---|
| baseline | 0.924 | 0.093 | 0.329 | 0.041 |
| +GLKAM | 0.932 | 0.089 | 0.316 | 0.039 |
| +GBPM | 0.931 | 0.090 | 0.319 | 0.039 |
| +GLKAM+GBPM | **0.934** | **0.088** | **0.314** | **0.038** |

*4.3.2. Effectiveness of GBPM*

**Table 3** show that the addition of GBPM leads to notable improvements in most evaluation metrics. The $\delta_1$ accuracy increases from 0.924 to 0.931, and the RMSE decreases from 0.322 to 0.319. Additionally, the $\log_{10}$ error drops from 0.041 to 0.039, indicating enhanced overall depth consistency. The improved global metrics suggest that GBPM enhances the model's capability to estimate large-scale depth variations. These gains stem from GBPM's global bin center prediction strategy, which avoids the limitations of locally adaptive binning and enables the model to reason about the entire scene's depth distribution more effectively.

As illustrated in **Figure 6**, the use of GBPM produces smoother depth transitions and more coherent global depth structures compared to the baseline. In particular, regions with large depth ranges—such as walls, floors, or open spaces—exhibit more continuous and geometrically plausible depth gradients. Although GBPM does not significantly refine fine structural edges on its own, it contributes substantially to the global consistency of the depth maps.

## 5. Conclusion and Future Work

In this paper, we propose a novel monocular depth estimation framework that effectively integrates both local contextual cues and global scene understanding. To achieve this, we introduce two key components: the Gated Large Kernel Attention Module (GLKAM), which captures multi-scale local features through large receptive field convolution and gated attention mechanisms; and the Global Bin Prediction Module (GBPM), which improves depth regression by predicting globally consistent bin centers rather than relying on local discretization. Extensive experiments conducted on NYU-V2 and KITTI datasets demonstrate that our method consistently outperforms existing approaches in both quantitative metrics and qualitative evaluations. Ablation studies further confirm that both GLKAM and GBPM individually contribute to performance gains, while their combination yields the best results by simultaneously enhancing local detail preservation and global depth consistency. Overall, our approach provides new insights into the design of depth estimation networks by emphasizing the importance of structured depth binning and multi-scale feature modeling. In future work, we plan to extend this framework to explore more efficient depth binning strategies and lightweight network designs for real-time applications.

**Acknowledgments**

This work was supported in part by the Natural Science Foundation of Shanghai (21ZR1421200), and part by the Shanghai Municipal Education Commission Artifcial Intelligence Program to Promote the Reform of Scientifc Research Paradigms and Empower Discipline Advancement (2024AI02010).